\documentclass[letterpaper, 10 pt, conference]{ieeeconf}  % Comment this line out if you need a4paper

\IEEEoverridecommandlockouts                              % This command is only needed if 
\let\labelindent\relax
\usepackage{enumitem}
\usepackage{booktabs}
\usepackage{caption}
\usepackage{xcolor}
\usepackage{hyperref}
\hypersetup{colorlinks=true}
\usepackage{mathtools}
\usepackage{amsfonts}
\usepackage{mathtools}
\usepackage{lipsum}
\usepackage{multicol}
\usepackage{multirow}
\usepackage{subcaption}
\usepackage{tcolorbox}

\usepackage{tabularx}
\usepackage{colortbl}

\usepackage[subtle,title=tight]{savetrees}

\definecolor{dgreen}{rgb}{0,0,0}
\definecolor{dyellow}{rgb}{.7,.7,0}
\definecolor{dred}{rgb}{1,0,0}
\definecolor{dblue}{rgb}{0,0,0.7}
\definecolor{dorange}{rgb}{0.9,0.5,0.1}
\definecolor{light-gray}{rgb}{0.8, 0.8, 0.8}
\definecolor{highlight}{HTML}{e3eeff}
\definecolor{comment-green}{rgb}{0.435, 0.576, 0.106}
\definecolor{prompt-gray}{HTML}{a7a7a7}
\definecolor{code-syntax}{HTML}{0060b1}

\newcommand{\ie}{i.e., }
\newcommand{\eg}{e.g., }

\newcommand{\command}[1]{\textcolor{comment-green}{#1}}
\newcommand{\prompt}[1]{\textcolor{prompt-gray}{#1}}

\usepackage{inconsolata}
\usepackage[T1]{fontenc}

\newcommand{\hlcode}[1]{\colorbox{highlight}{\makebox[0.96\linewidth][l]{#1}}}

\newcommand{\lmp}[1]{
\begin{tcolorbox}[boxsep=0pt,
                  left=3pt,
                  right=-4pt,
                  top=3pt,
                  bottom=3pt,
                  arc=0pt,
                  boxrule=0.5pt,
                  colframe=light-gray,
                  colback=white
                  ]
\small{  % potentially switch back to scriptsize if necessary
\ttfamily
#1
}
\end{tcolorbox}
}

\newcommand{\speciallmp}[1]{
\begin{tcolorbox}[
 enlarge top by=0.5em,
 boxsep=0pt,
                  left=3pt,
                  right=-4pt,
                  top=3pt,
                  bottom=3pt,
                  arc=0pt,
                  boxrule=0.5pt,
                  colframe=light-gray,
                  colback=white
                  ]
\small{  % potentially switch back to tiny if necessary
\ttfamily
#1
}
\end{tcolorbox}
}

\title{\LARGE \bf
\ \ \\
\ \ \\
Visual Language Maps for Robot Navigation
\vspace{0.75em}
}
\IEEEaftertitletext{\vspace{-1.25em}}
\author{Chenguang Huang$^{1}$, Oier Mees$^{1}$, Andy Zeng$^{2}$, Wolfram Burgard$^{3}$% <-this % stops a space
\thanks{$^{1}$University of Freiburg, Germany.}% <-this % stops a space
\thanks{$^{2}$Google Research, USA.}% <-this % stops a space
\thanks{$^{3}$University of Technology Nuremberg, Germany.}% <-this % stops a space
\thanks{This work has been supported partly by the German Federal Ministry of Education and Research under contract 01IS18040B-OML}% <-this % stops a space
\vspace{-3em}% <-this % stops a space
}
\date{}

\begin{document}
\maketitle
\thispagestyle{empty}
\pagestyle{empty}

%%%%%%%%%%%%%%%%%%%%%%%%%%%%%%%%%%%%%%%%%%%%%%%%%%%%%%%%%%%%%%%%%%%%%%%%%%%%%%%%
\begin{abstract}

Grounding language to the visual observations of a navigating agent can be performed using off-the-shelf visual-language models pretrained on Internet-scale data (\eg image captions). While this is useful for matching images to natural language descriptions of object goals, it remains disjoint from the process of mapping the environment, so that it lacks the spatial precision of classic geometric maps.
To address this problem, we propose VLMaps, a spatial map representation that directly fuses pretrained visual-language features with a 3D reconstruction of the physical world. VLMaps can be autonomously built from video feed on robots using standard exploration approaches and enables \emph{natural language indexing of the map} without additional labeled data.
Specifically, when combined with large language models (LLMs), VLMaps can be used to (i) translate natural language commands into a sequence of open-vocabulary navigation goals (which, beyond prior work, can be spatial by construction, \eg ``in between the sofa and the TV'' or ``three meters to the right of the chair'') directly localized in the map, and (ii) can be shared among multiple robots with different embodiments to generate new obstacle maps on-the-fly (by using a list of obstacle categories).
Extensive experiments carried out in simulated and real-world environments show that VLMaps enable navigation according to more complex language instructions than existing methods. Videos are available at \href{https://vlmaps.github.io}{https://vlmaps.github.io}.

\end{abstract}

%%%%%%%%%%%%%%%%%%%%%%%%%%%%%%%%%%%%%%%%%%%%%%%%%%%%%%%%%%%%%%%%%%%%%%%%%%%%%%%%
\section{INTRODUCTION}

People are excellent navigators of the physical world -- due in part to their remarkable ability to build cognitive maps~\cite{mcnamara1989subjective} that form the basis of spatial memory~\cite{chun1998contextual,newman2007learning} to (i) localize landmarks at varying ontological levels, such as a book; on the shelf; in the living room, or to (ii) determine whether the layout permits navigation between two points. Classic methods for robot navigation~\cite{thrun1998probabilistic, endres2012evaluation} build geometric maps for path planning and can parse goals from natural language commands~\cite{tellex2011understanding, macmahon2006walk}, but struggle to generalize to unseen instructions. Learning methods directly optimize for navigation policies grounded in language end-to-end (commands to actions)~\cite{anderson2018vision, anderson2021sim}, but require copious amounts of data.

Meanwhile, recent works show that visual-language models (VLMs)~\cite{radford2021learning,li2021language} pretrained on Internet-scale data (\eg image captions) can be used out-of-the-box to ground language to the visual observations of a navigating agent, without additional data collection or model fine-tuning. These models enable mobile robots to handle new instructions that specify unseen object goals and can be combined with exploration algorithms to search for the first instance of any object (CoW)~\cite{gadre2022clip} or traverse object-centric landmarks in graphs (LM-Nav)~\cite{shah2022lm}. While promising, these methods predominantly use VLMs as critics to match image observations to object goal descriptions, but do so in ways that remain disjoint from the mapping of the environment. Without grounding language onto a spatial representation, these systems may struggle to (i) recognize correspondences that associate independent observations of the same object, to (ii) localize spatial goals \eg ``in between the sofa and the TV'', or to (iii) build persistent representations that can be shared across different embodiments, \eg mobile robots, drones. Existing VLM-based solutions generalize to new object goals, but lose the spatial precision of classic geometric maps – is it possible to get the best of both?

 \begin{figure}[t]
    \vspace{0.5em}
	\centering
	\includegraphics[width=1\columnwidth]{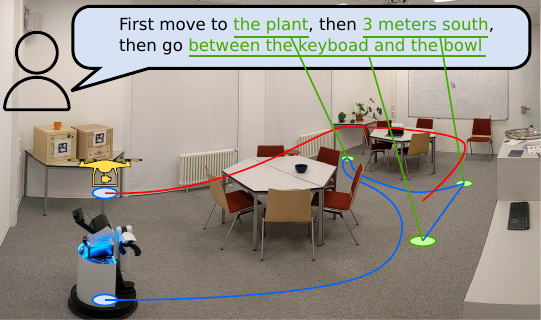}
	\caption{\small \textbf{VLMaps} is a spatial map representation in which pretrained visual-language model features are fused into a 3D reconstruction of the physical world. Spatially anchoring visual language features enables \emph{natural language indexing in the map}, which can be used to, \eg localize landmarks or spatial references with respect to landmarks -- enabling zero-shot spatial goal navigation without additional data collection or model finetuning.}
	\label{fig:cover_lady}
	\vspace{-1em}
\end{figure}

In this work, we investigate the utility of a \textit{spatial} visual-language map representation VLMaps, which fuses pretrained visual-language features from image observations directly with a 3D reconstruction of the physical world. VLMaps can be effectively built from video feed on robots using standard exploration algorithms. When paired with large language models (LLMs) in Socratic fashion~\cite{zeng2022socratic}, VLMaps can translate natural language instructions into a sequence of open-vocabulary goals, directly localized in the map. A key aspect of VLMaps is that they are spatial, which enables them to:
\begin{figure*}[t]
	\centering
	\includegraphics[width=\textwidth]{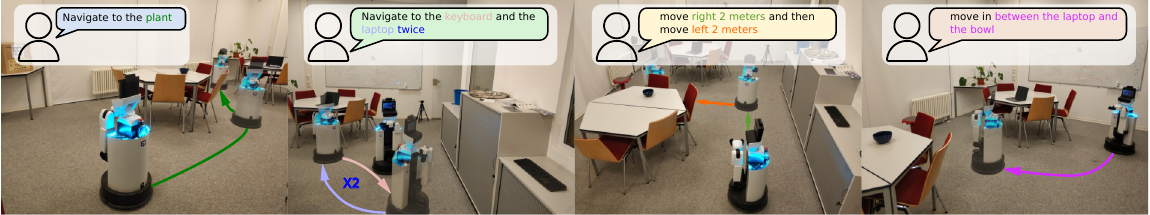}
	\caption{\small VLMaps enables a robot to perform complex zero-shot spatial goal navigation tasks given natural language commands, without additional data collection or model finetuning. }
	\label{fig:vlmap_applicatoins}
	\vspace{-1em}
\end{figure*}

\begin{itemize}[leftmargin=*]
  \item Localize spatial goals beyond object-centric ones, \eg ``in between the TV and sofa'' or ``to the right of the chair'' or ``kitchen area'' using code-writing LLMs, expanding beyond capabilities of CoW or LM-Nav.
  \item Generate new obstacle maps for new embodiments given natural language descriptions of landmark categories that they can or cannot traverse, \eg ``tables'' are obstacles for a large mobile robot, but traversable for a drone.
\end{itemize}

\noindent Extensive experiments show that using VLMaps enables more effective long-horizon multi-object goal navigation than baseline alternatives, \eg CoW~\cite{gadre2022clip} and LM-Nav~\cite{shah2022lm}, and, in particular, excels at enabling spatial open-vocabulary navigation tasks. We also provide ablations on different ways of constructing VLMaps with different language models as well as a discussion on limitations, which point to areas for future work. Code and videos are available at \href{https://vlmaps.github.io}{https://vlmaps.github.io}.

\section{Related Work}
\textbf{Semantic Mapping.}
The maturity of traditional SLAM techniques together with the advancements in semantic understanding capabilities of convolutional neural networks has recently spurred
considerable interest around augmenting 3D maps with semantic information~\cite{salas2013slam++, mccormac2017semanticfusion}. The literature has focused on either densely annotating 3D volumetric maps with 2D semantic segmentation CNNs~\cite{mccormac2017semanticfusion} or object-oriented approaches~\cite{runz2018maskfusion, mccormac2018fusion++, xu2019mid}, which build 3D maps around detected objects to enable object-level pose-graph optimization. Although progress has been made at generating more abstract maps, such as scene graphs~\cite{hughes2022hydra, wu2021scenegraphfusion}, current approaches are limited to a predefined set of semantic classes. In contrast to this, VLMaps are open-vocabulary semantic maps that, unlike prior work, enable \emph{natural language indexing in the map}.

\textbf{Vision and Language Navigation.} Recently, also Vision-and-Language Navigation (VLN) has received increased attention~\cite{anderson2018vision, krantz2020beyond}. Further work has focused on learning end-to-end policies that can follow route-based instructions on topological graphs of simulated environments~\cite{anderson2018vision, fried2018speaker, guhur2021airbert}. However, agents trained in this setting do not have low-level planning capabilities and rely heavily on the topological graph, limiting their real-world applicability~\cite{anderson2021sim}. Moreover, despite extensions to continuous state spaces~\cite{krantz2020beyond, krantz2021waypoint, hong2022bridging}, most of these learning-based methods are data-intensive. 

\textbf{Zero-shot Models.}
The recent success of large pretrained vision and language models~\cite{radford2021learning,brown2020language} has spurred a flurry of interest in applying their zero-sot capabilities to different domains including object detection and segmentation~\cite{kamath2021mdetr, gu2021open, li2021language}, robot manipulation~\cite{shridhar2022cliport, mees2022calvin, mees2022hulc, mees22hulc2}, and navigation~\cite{shah2022lm,gadre2022clip,chen2022nlmapsaycan}.
    Most related to our work is the approach denoted LM-Nav~\cite{shah2022lm}, which combines three pre-trained models to navigate via a topological graph in the real world. CoW~\cite{gadre2022clip} performs zero-shot language-based object navigation by combining CLIP-based~\cite{radford2021learning} saliency maps and traditional exploration methods. However, both LM-Nav \cite{shah2022lm} and CoW \cite{gadre2022clip} are limited to navigating to object landmarks and are less capable to understand finer-grained queries, such as ``to the left of the chair'' and ``in between the TV and the sofa''. In contrast, our method enables spatial language indexing beyond object-centric goals and can generate open-vocabulary obstacle maps. A concurrent work is NLMap~\cite{chen2022nlmapsaycan}, which demonstrates that VLMs can be used to build queryable scene representations to allow LLM robot planning~\cite{ahn2022can} with new objects and locations.

\section{Method}
\begin{figure*}[t]
	\centering
	\includegraphics[width=0.95\textwidth]{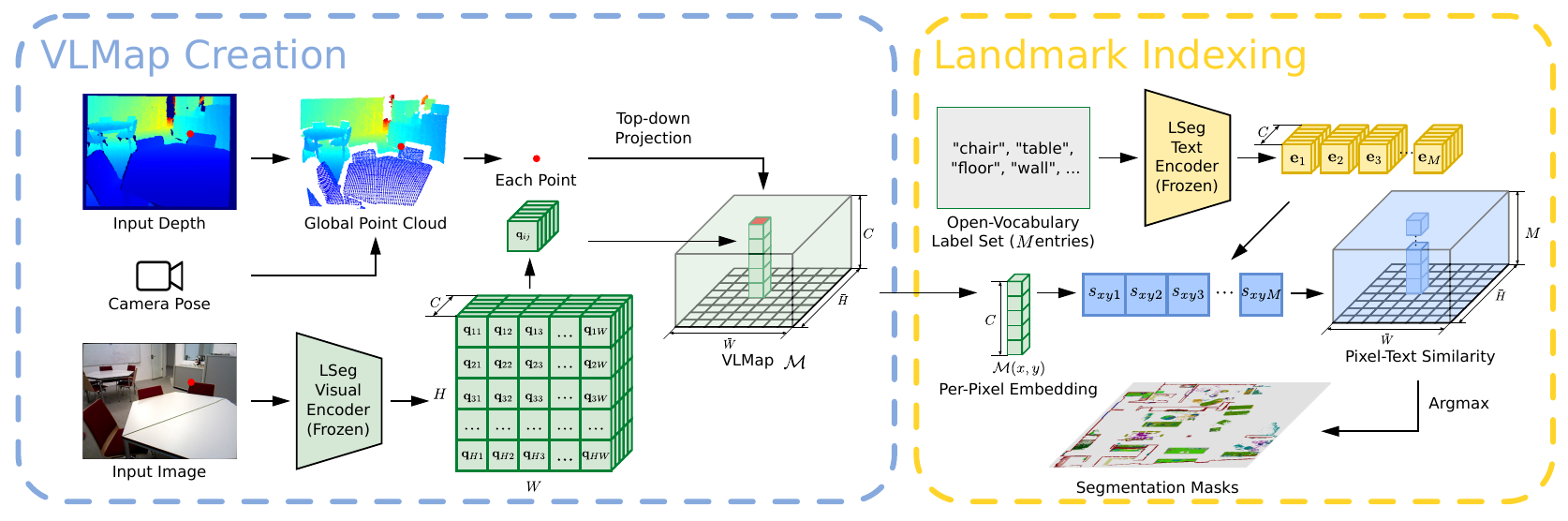}
	\caption{\small System overview. A VLMap is created by fusing pretrained visual-language features into the reconstruction of the environment to enable visual-spatial-language-based reasoning. By providing a list of open-vocabulary labels, we retrieve segmentation masks for semantic classes required by downstream applications.}
	\label{fig:system_overview}
	\vspace{-1em}
\end{figure*}

Our goal is to build a \textit{spatial} visual-language map representation, in which landmarks (``the sofa'') or spatial references (``between the sofa and the TV'') can be directly localized using natural language. We propose VLMaps as one such representation, which can be constructed using off-the-shelf visual-language models (VLMs) and standard 3D reconstruction libraries. In the following subsections, we describe (i) how to build a VLMap (Sec.~\ref{sec:sub_map_creation}), (ii) how to use these maps to localize open-vocabulary landmarks (Sec.~\ref{sec:sub_landmark_indexing}), (iii) how to build open-vocabulary obstacle maps from a list of obstacle categories for different robot embodiments (Sec.~\ref{sec:obstacle_maps}), and (iv) how VLMaps can be used together with large language models (LLMs) for zero-shot spatial goal navigation on real robots from natural language commands (Sec.~\ref{sec:sub_language_spatial_navigation}), without additional data collection or model fine-tuning. Our pipeline is visualized in Fig.~\ref{fig:system_overview}. 

\subsection{Building a Visual-Language Map}
\label{sec:sub_map_creation}
The key idea behind VLMaps is to fuse pretrained visual-language features with a 3D reconstruction. 
We achieve this by computing dense pixel-level embeddings from an existing visual-language model (over the video feed of the robot) and by back-projecting them onto the 3D surface of the environment (captured from depth data used for reconstruction with visual odometry).

In our work, we utilize LSeg~\cite{li2021language} as the visual-language model, a language-driven semantic segmentation model that segments the RGB images based on a set of free-form language categories. The LSeg visual encoder maps an image such that the embedding of each pixel lies in the CLIP feature space. In our approach, wie fuse the LSeg pixel embeddings with their corresponding 3D map locations. In this way, without explicit manual segmentation labels, we incorporate a powerful language-driven semantic prior that inherits the generalization capabilities of VLMs. The only assumption we make is access to odometry, which is readily available from RGB-D SLAM systems and enables us to build a map from sequences of RGB-D images, 

Formally, we define VLMap as $\mathcal{M} \in \mathbb{R}^{\bar{H} \times \bar{W} \times C}$, where $\bar{H}$ and $\bar{W}$ represent the size of the top-down grid map, and $C$ represents the length of the VLM embedding vector for each grid cell. Together with the scale parameter $s$, a VLMap $\mathcal{M}$ represents an area with size $s\bar{H} \times s\bar{W}$ meters.
To build the map, we, for each RGB-D frame, back-project all the depth pixels $\mathbf{u} = (u, v)$ to form a local depth point cloud that we transform to the world frame, $\mathbf{P}_{k} = D(\mathbf{u}) K^{-1} \mathbf{\tilde{u}}$ and $\mathbf{P}_{W} = T_{Wk} \mathbf{P}_{k}$
% \begin{equation}
% \mathbf{P}_{k} = D(\mathbf{u}) K^{-1} \mathbf{\tilde{u}}
% \end{equation}
% \begin{equation}
% \mathbf{P}_{W} = T_{Wk} \mathbf{P}_{k}
% \end{equation}
where $\mathbf{\tilde{u}} = (u, v, 1)$, $K$ is the intrinsic matrix of the depth camera, $D(\mathbf{u}) \in \mathbb{R}$ is the depth value of the pixel $\mathbf{u}$, $T_{Wk}$ is the transformation from the world coordinate frame to the k-th camera frame, $\mathbf{P}_{k} \in \mathbb{R}^{3}$ is the 3D point position in the k-th frame, and $\mathbf{P}_{W} \in \mathbb{R}^{3}$ is the 3D point position in the world coordinate frame. We then project the point $\mathbf{P}_{W}$ to the ground plane and get the pixel $\mathbf{u}$'s corresponding position on the grid map,
\begin{equation}
\label{eq:px_map}
p^{x}_{map} = \Bigl\lfloor \frac{\bar{H}}{2} + \frac{P^{x}_{W}}{s} + 0.5 \Bigr\rfloor,~ p^{y}_{map} = \Bigl\lfloor \frac{\bar{W}}{2} - \frac{P^{z}_{W}}{s} +0.5 \Bigr\rfloor
\end{equation}
where $p^{x}_{map}$ and $p^{y}_{map}$ represent the coordinates of the projected point in the map $\mathcal{M}$.

Once we build the grid map, we apply LSeg's visual encoder $f(\mathcal{I}): \mathbb{R}^{H \times W \times 3} \rightarrow \mathbb{R}^{H \times W \times C}$ to the RGB image $\mathcal{I}_{k}$ and generate the pixel-level embedding $\mathcal{F}_{k} \in \mathbb{R}^{H \times W \times C}$. Given the RGB-D registration, we project each image pixel $\mathbf{u}$'s embedding $\mathbf{q} = \mathcal{F}_{k}(\mathbf{u}) \in \mathbb{R}^{C}$ to its corresponding grid cell location $(p^{x}_{map}, p^{y}_{map})$ in the top-down grid map.  Intuitively, there exist multiple 3D points projecting to the same grid location in the map. Thus, we average their embeddings, $\mathcal{M} ( p^{x}_{map},\ p^{y}_{map} ) = \frac{1}{n}  \sum_{i=1}^{n} \mathbf{q}_i$
% \begin{equation}
% \mathcal{M} ( p^{x}_{map},\ p^{y}_{map} ) = \frac{1}{n}  \sum_{i=1}^{n} \mathbf{q}_i
% \end{equation}
where $\mathcal{M} ( p^{x}_{map},\ p^{y}_{map} ) \in \mathbb{R}^{C}$ represents the map features at the grid position $(p^{x}_{map}, p^{y}_{map})$, $n$ represents the total number of points projecting to the grid location $(p^{x}_{map},\ p^{y}_{map})$, and $\mathbf{q}_i \in \mathbb{R}^{C}$ denotes the corresponding pixel embedding of each point. We note that these $n$ points might not only come from a single frame, but also from points from multiple frames. Therefore, the resulting features contain the averaged embeddings from multiple views of the same object. 

\subsection{Localizing Open-Vocabulary Landmarks}
\label{sec:sub_landmark_indexing}
We now describe how to localize landmarks in VLMaps with free-form natural language.
Formally, we define the input language list as $\mathcal{L} = [\mathbf{l}_0, \mathbf{l}_1, \ldots, \mathbf{l}_M]$ where $\mathbf{l}_i$ represents the i-th category in text form, and $M$ represents the number of categories defined by the user. Some examples of the input language list are [``chair'', ``sofa'', ``table'', ``other''] or [``furniture'', ``floor'', ``other'']. As Li \emph{et al.} \cite{li2021language}, we apply the pre-trained CLIP text encoder \cite{radford2021learning} to convert such list of texts into a list of vector embeddings $[\mathbf{e}_0, \mathbf{e}_1, \ldots, \mathbf{e}_M],\ \mathbf{e} \in \mathbb{R}^{C}$, which are organized into an embedding matrix $E \in \mathbb{R}^{M \times C}$, where each row of the matrix represents the embedding of a category. The map embeddings $\mathcal{M}$ are also flattened into a matrix $Q \in \mathbb{R}^{\bar{H}\bar{W} \times C}$, where each row represents the embedding of a pixel in the top-down grid map. We then compute the pixel-to-category similarity matrix $S = Q \cdot E^{T}$,
% \begin{equation}
% S = Q \cdot E^{T}
% \end{equation}
where $S \in \mathbb{R}^{\bar{H}\bar{W} \times M}$. Each element $S_{ij}$ in the matrix stores the similarity value between a pixel and a text category, indicating how likely this pixel belongs to the class. By applying the $\mathrm{argmax}$ operator along the row direction to $S$ and reshaping the resulting vector to shape $\bar{H} \times \bar{W}$, we get the final segmentation result $R \in \mathbb{R}^{\bar{H} \times \bar{W}}$. Each element $R_{ij}$ represents the label index of the input language list $\mathcal{L}$ at the grid map location $(i, j)$. With the final resulting matrix $R$, we compute the most related language-based category for every pixel in the grid map. 

\subsection{Generating Open-Vocabulary Obstacle Maps}
\label{sec:obstacle_maps}
Building a VLMap enables us to generate obstacle maps that inherit the open-vocabulary nature of the VLMs used (LSeg and CLIP). Specifically, given a list of obstacle categories described with natural language, we can localize those obstacles at runtime to generate a binary map for collision avoidance and/or shortest path planning. A prominent use case for this is sharing a VLMap of the same environment between different robots with different embodiments (\ie cross-embodiment problem \cite{zakka2022xirl,ganapathi2022implicit}), which may be useful for multi-agent coordination \cite{wu2021spatial}. For example, a large mobile robot may need to navigate around a table (or other large furniture), while a drone can directly fly over it. By simply providing two different lists of obstacle categories -- one for the large mobile robot (that contains ``table''), and another for the drone (that does not), we can generate two distinct obstacles maps for the two robots to use respectively, sourced on-the-fly from the same VLMap.

To do so, we first extract an obstacle map $\mathcal{O} \in \{0, 1\}^{\bar{H} \times \bar{W}}$ where each projected position of the depth point cloud in the top-down map is assigned~1, and otherwise~0. To avoid points from the floor or the ceiling, points $P_{W}$ are filtered out depending on their height,
\begin{equation}
\mathcal{O}_{ij} = 
    \begin{dcases}
        1,& t_1 \leq P^{y}_{W} \leq t_2\ and\ p^{x}_{map} = i\ and\ p^{y}_{map} = j\\
        0,& \text{otherwise}
    \end{dcases}
\end{equation}
where $t_1, t_2 \in \mathbb{R}$ are the lower and upper thresholds for the $y$-component of the point $P_{W}$. 
Second, to obtain obstacle maps tailored to a certain embodiment, we define a list of potential obstacle categories $\mathcal{L}_{obs} = [\mathbf{l}_{obs0}, \mathbf{l}_{obs1}, \ldots, \mathbf{l}_{obsM}]$, where $\mathbf{l}_{obsi}$ represents the i-th obstacle category in language, and $M$ represents the total number of obstacle categories defined by the user. We then apply the open-vocabulary landmark indexing introduced in Sec.~\ref{sec:sub_landmark_indexing} and obtain segmentation masks for all defined obstacles. For a specific embodiment $k$, we choose a subset of classes out of the whole potential obstacle list $\mathcal{L}_{obs}$ and take the union of their segmentation masks to get the obstacles mask $\tilde{\mathcal{O}}_{em_k}$. We ignore false predictions of obstacles on floor region in $\tilde{\mathcal{O}}_{em_k}$ by taking the intersection with $\mathcal{O}$ to get the final obstacle map $\mathcal{O}_{em_k}$.

\subsection{Zero-Shot Spatial Goal Navigation from Language}
\label{sec:sub_language_spatial_navigation}
In this section, we describe our approach to long-horizon (spatial) goal navigation, given a set of landmark descriptions specified by natural language instructions such as
\lmp{
\prompt{
move first to the left side of the counter, then \\
move between the sink and the oven, then move back \\
and forth to the sofa and the table twice
% move first to the left side of the counter in front \\
% of you, face the counter and then move to the west \\
% of the counter, later, with the counter on your \\
% right, go to the east of the chair in front of you, \\
% and finally move to the sofa in front of you
}
}
\noindent Notably different from prior work~\cite{gadre2022clip, shah2022lm}, VLMaps allow us to reference precise spatial goals such as: ``in between the sofa at the TV'' or ``three meters to the east of the chair.''
Specifically, we use a large language model (LLM) to interpret the input natural language commands and break them down into subgoals~\cite{ahn2022can,shah2022lm,zeng2022socratic}. In contrast to prior work, which may reference these subgoals with language and map to low-level policies with semantic translation~\cite{huang2022language} or affordances~\cite{ahn2022can,huang2022inner,zeng2019learning, borja22icra}, we leverage the code-writing capabilities of LLMs to generate executable Python robot code~\cite{liang2022code,mees22hulc2,chen2021evaluating,brown2020language} that can (i) make precise calls to parameterized navigation primitives, and (ii) perform arithmetic when needed. The generated code can directly be executed on the robot with the built-in Python \textbf{exec} function.

Note that recent works~\cite{liang2022code,mees22hulc2,chen2021evaluating,brown2020language} have shown that code-writing language models (\eg Codex~\cite{chen2021evaluating}) trained on billions of lines of code from Github can be used to synthesize new simple Python programs from docstrings. In this work, we re-purpose these models for mobile robot planning, by priming them with several input examples of natural language commands (formatted as comments) paired with corresponding robot code (via few-shot prompting). The robot code can express functions or logic structures (if-then-else statements or for/while loops) and parameterize API calls (\eg robot.move\_to(target\_name) or robot.turn(degrees). The full list is available in the Appendix, Sec.~\ref{sec:navigation-primitives}) that map to spatial behaviors specified by the language commands. At test time, the models can subsequently take in new commands and autonomously re-compose API calls to generate new robot code respectively (prompt in {\color{prompt-gray}gray}, input task commands in \command{green}, and generated outputs are \colorbox{highlight}{highlighted}):
\lmp{
\prompt{
\# move a bit to the right of the fridge\\
robot.move\_to\_right(`refrigerator')\\
% \# move in between the couch and bookshelf\\
% robot.move\_in\_between(`couch', `bookshelf')\\
\# face the toilet\\
robot.face(`toilet')\\
\# move to the west of the chair\\
robot.move\_west(`chair')\\
\# turn right 20 degrees\\
robot.turn(20)\\
% \# find any chairs in the environment\\
% robot.move\_to\_object(`chair')\\
% \# with the television on your left\\
% robot.with\_object\_on\_left(`television')\\
% \# move right 2 meters\\
% robot.turn(90)\\
% robot.move\_forward(2)\\
\# move back and forth to the chair and table 3 times\\
pos1 = robot.get\_pos(`chair')\\
% pos2 = robot.get\_pos(`table')\\
% for i in range(3):\\
% \hspace*{4mm}robot.move\_to(pos1)\\
% \hspace*{4mm}robot.move\_to(pos2)\\
% \# move 3 meters south of the chair\\
% robot.move\_south(`chair')\\
% robot.face(`chair')\\
% robot.turn(180)\\
% robot.move\_forward(3)
...\\
\# move forward for 3 meters\\
robot.move\_forward(3)
}}
\speciallmp{
\command{\# move first to the left side of the counter, then \\
move between the sink and the oven, then move back and \\
forth to the sofa and the table twice}\\
\hlcode{robot.move\_to\_left(`counter')}\\
\hlcode{robot.move\_in\_between(`sink', `oven')}\\
\hlcode{pos1 = robot.get\_pos(`sofa')}\\
\hlcode{pos2 = robot.get\_pos(`table')}\\
\hlcode{for i in range(2):}\\
\hlcode{\hspace*{4mm}robot.move\_to(pos1)}\\
\hlcode{\hspace*{4mm}robot.move\_to(pos2)}\\
\command{\# move 2 meters north of the laptop, then move 3 \\
meters rightward}\\
\hlcode{robot.move\_north(`laptop')}
\hlcode{robot.face(`laptop')}
\hlcode{robot.turn(180)}
\hlcode{robot.move\_forward(2)}
\hlcode{robot.turn(90)}
\hlcode{robot.move\_forward(3)}
}

The code-writing LLM generates code that not only references the new landmarks mentioned in the language commands (as comments), but also can chain together new sequences of API calls to follow unseen instructions accordingly. The prompt has been truncated for brevity here. Please see the full prompt in the Appendix (Sec.~\ref{sec:full-prompts}).

The navigation primitive functions being called by the language model (\eg robot.move\_to\_left(`counter')) use a pre-generated VLMap to localize the coordinates of the open-vocabulary landmarks (``counter'') in the maps (described in Sec.~\ref{sec:sub_landmark_indexing}) modified with predefined scripted offsets (to define ``left''). We then navigate to these coordinates using an off-the-shelf navigation stack that takes as input the embodiment-specific obstacle map (generated using the same VLMap, with the process described in Sec.~\ref{sec:obstacle_maps}).

\section{Experiments}
The goals of our experiments are four-fold:
(i) to quantitatively evaluate our VLMaps approach against recent open-vocabulary navigation baselines on the standard task of multi-object goal navigation (Sec.~\ref{sec:exp_multi_object_navigation}),
(ii) to investigate whether our method can better navigate to \textit{spatial} goals specified by language commands versus alternative approaches (Sec.~\ref{sec:exp_spatial_goal_navigation_language}),
(iii) to study whether VLMaps with their capacity to specify open-vocabulary obstacle maps can provide utility in improving the navigation efficiency of different robots with different embodiments (Sec.~\ref{sec:exp_multi_embodiment_navigation}),
and (iv) to demonstrate on real robots that VLMaps can enable zero-shot spatial goal navigation given unseen language instructions (Sec.~\ref{sec:exp_real_world}).

\subsection{Simulation Setup}
\label{sec:sim_experiments}
\noindent\textbf{Experimental setup.} We use the Habitat simulator~\cite{habitat19iccv} with the Matterport3D dataset~\cite{Matterport3D} for the evaluation of multi-object and spatial goal navigation tasks. The dataset contains a large set of realistic indoor scenes that help evaluate the generalization capabilities of navigating agents. To evaluate the creation of open-vocabulary multi-embodiment obstacle maps, we adopt the AI2THOR simulator due to its support of multiple agent types, such as LoCoBot and drone. In these two environments, the robot is required to navigate in a continuous environment with actions: \textbf{move forward 0.05 meters}, \textbf{turn left 1 degree}, \textbf{turn right 1 degree} and \textbf{stop}. For map creation in Habitat, we collect 12,096 RGB-D frames across ten different scenes and record the camera pose of each frame. Similarly, we collect 1,826 RGB-D frames across ten rooms in AI2THOR.
 \begin{figure}[t]
    \begin{subfigure}{.185\textwidth}
      \centering
      % include third image
      \includegraphics[width=\linewidth]{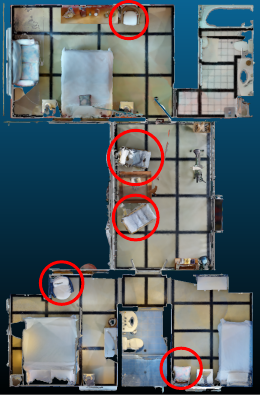}  
      \caption{Top-Down Map}
      \label{fig:object_localization_top_down}
    \end{subfigure}
    \hfill
    \begin{subfigure}{.6\textwidth}
        \begin{subfigure}{.23\textwidth}
          \centering
          % include third image
          \includegraphics[width=\linewidth]{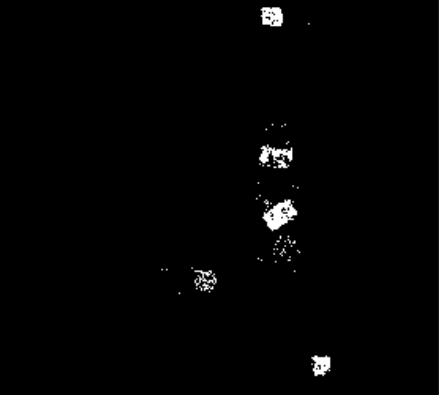}  
          \caption{GT}
          \label{fig:object_localization_gt}
        \end{subfigure}
        \begin{subfigure}{.23\textwidth}
          \centering
          % include third image
          \includegraphics[width=\linewidth]{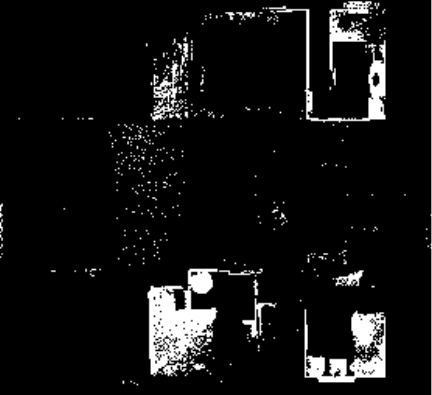}  
          \caption{CLIP Map}
          \label{fig:object_localization_clip_map}
        \end{subfigure}
        \vfill
        \begin{subfigure}{.23\textwidth}
          \centering
          % include third image
          \includegraphics[width=\linewidth]{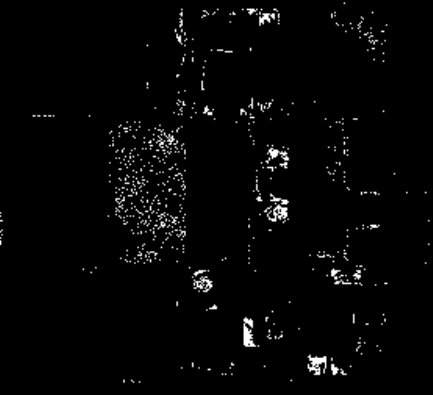}  
          \caption{CoW}
          \label{fig:object_localization_cow}
        \end{subfigure}
        \begin{subfigure}{.23\textwidth}
          \centering
          % include third image
          \includegraphics[width=\linewidth]{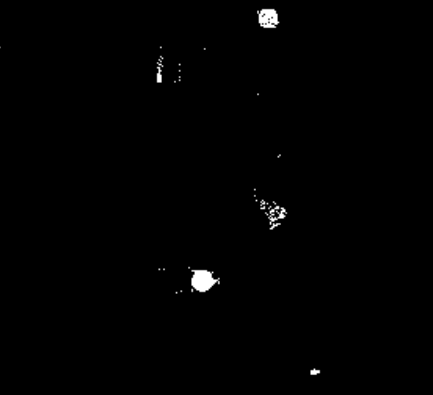}  
          \caption{VLMaps}
          \label{fig:object_localization_vlmaps}
        \end{subfigure}
    \end{subfigure}

	\caption{\small Object mask for object type ``chair''. \ref{fig:object_localization_top_down} shows the top-down map of the scene and the red circles specify the locations of type ``chair''. \ref{fig:object_localization_gt} shows the ground truth mask for type ``chair'' and \ref{fig:object_localization_clip_map}, \ref{fig:object_localization_cow}, \ref{fig:object_localization_vlmaps} show the predicted masks by CLIP Map, CoW and VLMaps.  % \TODO{figures format and alignment, it is hard coded now
	}
	\label{fig:object_localization_chair}
	\vspace{-1em}
\end{figure}

\noindent\textbf{Baselines.} We evaluate VLMaps against three baseline methods, all of which utilize visual-language models and are capable of zero-shot language-based navigation:
\begin{itemize}[leftmargin=*]
  \item LM-Nav~\cite{shah2022lm} creates a graph where image observations of an environment are stored as nodes while the proximity between images are represented as edges. By combining GPT-3 and CLIP, it parses language instructions into a list of landmarks and plans on the graph towards corresponding nodes. 
  \item CLIP on Wheels (CoW)~\cite{gadre2022clip} achieves language-based object navigation by building a saliency map for the target category with CLIP and GradCAM~\cite{selvaraju2020grad}. By thresholding the saliency values, it retrieves a segmentation mask for the target object category and then plans the path on the map.
  \item CLIP-features-based map (CLIP Map) is an ablative baseline that generates a feature map for the environment in a similar way as ours. Instead of using LSeg visual features, it projects the CLIP visual features onto the map averaged across views. Object category masks are generated by thresholding the similarity between map features and the object category features.
\end{itemize}
For additional context and analysis, we also report results from a system that has access to a ground truth semantic map for navigation, to provide a systems-level upper bound on performance.

\subsection{Multi-Object Navigation}
\label{sec:exp_multi_object_navigation}
We collect 91 sequences of tasks for the evaluation of object navigation. In each sequence, we randomly specify a starting position of the robot in one scene and then pick four among 30 object categories as subgoal object types. The robot is required to navigate to these four subgoals sequentially. In each sequence of subgoals, when the robot reaches one subgoal category, it should call the \textbf{stop} action to indicate its progress. We consider the navigation to one subgoal as success when the distance of stop position from the correct object is within one meter. To evaluate the long-horizon navigation capabilities of the agents, we compute the success rate (SR) of continuously reaching one to four subgoals in a sequence, shown in Tab.~\ref{table:multi_obj_nav}. We also report the independent subgoal success rate, which indicates the total successful subgoals number divided by the total subgoals number (364 subgoals). % In the evaluation, we assume that the robot has access to it position and the orientation in the environment.

\begin{table}[h]
  \setlength\tabcolsep{5.2pt}
  \centering
  \begin{tabular}{lccccc}
  \toprule
\multirow{2}[1]{*}{Tasks}  & \multicolumn{4}{l}{No. Subgoals in a Row} & Independent      \\                
                    \cmidrule(lr){2-5}
        &   1             & 2           & 3           & 4    &  Subgoals\\
\midrule
LM-Nav \cite{shah2022lm}& 26                  & 4               & 1               & 1               & 26         \\
CoW \cite{gadre2022clip}& 42                  & 15               & 7               & 3               & 36         \\
CLIP Map        & 33                  & 8                & 2               & 0               & 30         \\
VLMaps (ours) & \textbf{59}         & \textbf{34}      & \textbf{22}     & \textbf{15}     & \textbf{59}     \\
\midrule
GT Map    & 91         & 78      & 71     & 67     & 85          \\
  \bottomrule
  \end{tabular}
  \caption{The VLMaps-approach performs favorably over alternative open-vocabulary baselines on multi-object navigation (success rate [\%]) and specifically excels on longer-horizon tasks with multiple sub-goals.}
  \label{table:multi_obj_nav}
  \vspace{-1em}
\end{table}

We observe that VLMaps performs consistently better compared to all baselines. LM-Nav has a weak performance as it is only able to navigate to locations represented by images stored in graph nodes. To obtain more insights into the map-based methods, we visualize the object masks generated by VLMaps, CoW, and CLIP Map, in comparison to GT, in Fig.~\ref{fig:object_localization_chair}. The masks generated by CoW (Fig.~\ref{fig:object_localization_cow}) and CLIP (Fig.~\ref{fig:object_localization_clip_map}) both contain considerable false positive predictions. Since the planning generates the path to the nearest masked target area, these predictions lead to planning towards wrong goals. In contrast, the predictions obtained with VLMaps shown in Fig.~\ref{fig:object_localization_vlmaps} are less noisy, which leads to higher success rates in object navigation.

\subsection{Zero-Shot Spatial Goal Navigation from Language}
\label{sec:exp_spatial_goal_navigation_language}
In these experiments, we investigate the performance of VLMaps versus other baselines for zero-shot \textit{spatial} goal navigation from language. Our benchmark consists of 21 trajectories in seven scenes, with manually specified corresponding language instructions for evaluation. Each trajectory contains four different spatial locations as subgoals. Examples of subgoals are ``east of the table'', ``in between the chair and the sofa'', or ``move forward 3 meters''. There are also instructions for the robot to realign itself in reference to nearby objects such as ``with the counter on your right''. We only consider a subgoal as having been achieved, when the robot reaches the subgoal location within a range of one meter. We compute the in-a-row success rate in the same way as in Sec.~\ref{sec:exp_multi_object_navigation}. For all map-based methods, including CoW, CLIP Map, ground truth semantic map and our method, we apply the code generation techniques introduced in Sec.~\ref{sec:sub_language_spatial_navigation}. For LM-Nav, we simply use the same parsing method in the original paper~\cite{shah2022lm} to break down the language instruction into subgoals.

\begin{table}[h]
  \setlength\tabcolsep{5.2pt}
  \centering
  \begin{tabular}{lcccc}
  \toprule
\multirow{2}[1]{*}{Tasks}  & \multicolumn{4}{l}{No. Subgoals in a Row}       \\                
                    \cmidrule(lr){2-5}
                   & 1             & 2           & 3           & 4             \\
\midrule
LM-Nav \cite{shah2022lm}& 5                  & 5               & 0               & 0                        \\
CoW \cite{gadre2022clip}& 33                  & 5               & 0               & 0                        \\
CLIP Map           & 19                  & 0                & 0               & 0                        \\
VLMaps (ours)    & \textbf{62}         & \textbf{33}      & \textbf{14}     & \textbf{10}          \\
\midrule
GT Map    & 76        & 48      & 33     & 29          \\

  \bottomrule
  \end{tabular}
  \caption{The VLMaps approach can navigate to spatial goals specified by natural language and outperforms other open-vocabulary zero-shot navigation baseline alternatives (success rate [\%]) in this setting.}
  \label{table:spatial_lang_nav}
  \vspace{-1em}
\end{table}

Tab.~\ref{table:spatial_lang_nav} summarizes the zero-shot spatial goal navigation success rates. Our method outperforms other baselines in this task. Different from object navigation tasks where agents only need to approach a certain object type within a range disregarding the relative spatial shift to the object, the language-based spatial goal navigation tasks require the robot to accurately arrive at the described location in reference to the object. This poses a bigger challenge to the landmark localization ability of the method. The low localization ability of CoW and CLIP Map analyzed in the previous section (Sec.~\ref{sec:exp_multi_object_navigation}) leads to their high failure rates in this task.

\subsection{Cross-Embodiment Navigation}
\label{sec:exp_multi_embodiment_navigation}
We study the ability of VLMaps to improve navigation efficiency by retrieving different obstacle maps for navigation with different embodiments (given the same VLMap). We evaluate more than 100 sequences of subgoals as in Sec.~\ref{sec:exp_multi_object_navigation} in the AI2THOR simulator. We evaluate VLMaps on both a LoCoBot and a drone to test its capability of generating obstacle maps at runtime for multi-embodiment navigation. We apply the open-vocabulary obstacle map generation method in Sec.~\ref{sec:obstacle_maps} to create an obstacle map for the drone (drone map) and one for the LoCoBot (ground map) by defining obstacles for them differently (see the prompts in Appendix Sec.~\ref{sec:prompt_obs_map}). We test the navigation ability of these embodiments with three setups: a LoCoBot with a ground map, a drone with a ground map, and a drone with a drone map.

We evaluate the Success Rate (SR) and the Success rate weighted by the (normalized inverse) Path Length (SPL) \cite{anderson2018evaluation} defined as: $SPL = \frac{1}{N} \sum_{i=1} ^{N} S_i \frac{\textit{l}_i}{max(\textit{p}_i, \textit{l}_i)}$
% \begin{equation}
% SPL = \frac{1}{N} \sum_{i=1} ^{N} S_i \frac{\textit{l}_i}{max(\textit{p}_i, \textit{l}_i)}
% \end{equation}
where $N$ is the total number of evaluated tasks, $S_i \in \{0, 1\}$ is the binary indicator of success, $\textit{l}_i$ denotes the ground truth shortest path length, and $\textit{p}_i$ denotes the actual path length of the agent in navigation. This metric indicates how efficient the actual path is compared to the ground truth shortest path when the navigation task is achieved. In our three setups, the ground truth trajectories for the LoCoBot and the drone are planned on floor-level and on height level of 1.7 meters respectively.

\begin{table}[h]
  \setlength\tabcolsep{1pt}
  \centering
  \begin{tabular}{lccccccccc}
  \toprule
\multirow{3}[3]{*}{Tasks}  & \multicolumn{8}{c}{No. Subgoals in a Row} &  Independent      \\                
                    \cmidrule(lr){2-9}
        & \multicolumn{2}{c}{1}  & \multicolumn{2}{c}{2} & \multicolumn{2}{c}{3} & \multicolumn{2}{c}{4} &  Subgoals\\
         \cmidrule(lr){2-3}  \cmidrule(lr){4-5}  \cmidrule(lr){6-7}  \cmidrule(lr){8-9}  \cmidrule(lr){10-10}
        & SR      & SPL      & SR      & SPL      & SR      & SPL      & SR      & SPL      & SR      \\
\midrule
LoCoBot (ground map)     & 53      & \textbf{49.0}      & 28      & \textbf{17.8}       & 14      & 6.7       & 6       & \textbf{2.5}        & 52.3      \\
Drone (ground map)       & 53      & 41.8       & 28      & 15.5       & 14      & 5.3       & 6       & 2.0        & 53.3      \\
Drone (drone map)        & \textbf{56} & 45.4 & \textbf{30} & 16.3 & \textbf{17} & \textbf{7.0} & \textbf{7} & \textbf{2.5} & \textbf{55.0}      \\

  \bottomrule
  \end{tabular}
  \caption{VLMaps generate different obstacle maps for different robot embodiments, conditioned on a list of obstacle categories. This improves object navigation efficiency (Success [\%] weighted by Path Length, SPL).}
  \label{table:multi_embodiment_nav}
  \vspace{-1em}
\end{table}

\begin{figure}[t]
	\centering
	\includegraphics[width=1\columnwidth]{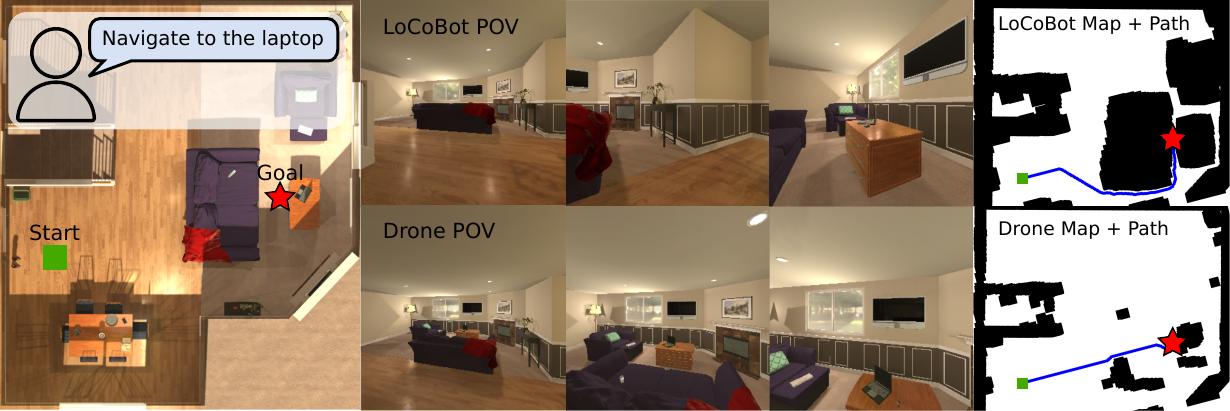}
	\caption{\small VLMaps enable different embodiments to define their own obstacle maps for navigation. The left image shows the top-down view of an environment. The middle columns show the observations of agents during navigation. The images on the right demonstrate the obstacles maps generated for different embodiments and the corresponding navigation paths.} 
	\label{fig:multi_embodiment}
	\vspace{-1em}
\end{figure}

The results provided in Tab.~\ref{table:multi_embodiment_nav} show that the average navigation success rates of the ground-map version of the LoCoBot and the drone are similar because the same obstacles map is used for planning. However, there is an obvious gap between their SPL values. This is because when the drone does not have access to a customized obstacle map, it fails to benefit from flying over ground objects to improve the navigation efficiency. In contrast, while achieving similar success rate compared to the drone with a ground map, the drone with a drone map manages to navigate with higher path efficiency, reflected by the increased SPL values. The comparable SPL values for the drone with the drone map and the LoCoBot with the ground map shows that VLMaps help to generalize the navigation efficiency among different embodiments. An example of the multi-embodiment object navigation task is shown in Fig.~\ref{fig:multi_embodiment}, where by defining a more efficient obstacles map, the drone flies over the sofa and reaches the laptop target directly, while the LoCoBot has to move aside first to avoid colliding with the sofa.

\subsection{Real Robot Experiments}
\label{sec:exp_real_world}

We also perform real-world experiments using the HSR mobile robot for indoor navigation given natural language commands. For map creation, we record 374 frames for the evaluated scene and use an off-the-shelf RGB-D SLAM solution, RTAB-Map~\cite{labbe2019rtab} to estimate the camera poses. During inference, we also use the global localization module of RTAB-Map to initialize the robot pose. We test our VLMaps in a semantically rich indoor scene with more than ten different classes of objects. We define 20 different language-based spatial goals for testing purposes. Across different test runs, we initialize the robot at different locations. 

The robot finishes ten navigation goals out of the 20. Among the successful trials, six of them are spatial goals like ``move between the chair and the wooden box'' or ``move to the south of the table''. three of them are goals relative to the current position of the robot like ``move 3 meters right and then move 2 meters left''. Another one is an instruction with repetition: ``move between the keyboard and the laptop twice''. We observe that failure cases are caused by: 1) inaccurate depth, which introduces noise during the map creation and decreases the landmark indexing accuracy and 2) action noise, which can negatively influence the navigation performance at test time. Overall, these results demonstrate the ability of VLMaps to index landmarks with natural language in the real world and, more importantly, its applicability to achieve a wide variety of open-vocabulary language-based spatial navigation goals.

\section{Discussion and Limitations}
% \section{Conclusion} <- I think we can directly conclude with discussion section
In this work, we propose VLMaps, a spatial map representation enriched with pretrained visual-language features, which enables natural language indexing in the map. When combined with large language models, VLMaps can be applied in zero-shot spatial goal navigation and can be shared among multiple robots with different embodiments to generate new obstacles map in runtime. VLMaps are not without limitations. Notably, they remain sensitive to 3D reconstruction noise and odometry drift during navigation. They also cannot resolve object ambiguities during landmark indexing when the scene is cluttered with similar objects. In future work, we plan to improve VLMaps with better visual language models and to extend it to scenes with dynamic objects and moving humans.

%extending it to a language-based active SLAM system. 
% Overall, we are excited by the confluence of pre-trained visuo-lingual models and embodied agents towards scaling robot learning.
%Another potential future work is to enable language-based landmark indexing at instance level.

%%%%%%%%%%%%%%%%%%%%%%%%%%%%%%%%%%%%%%%%%%%%%%%%%%%%%%%%%%%%%%%%%%%%%%%%%%%%%%%%

%%%%%%%%%%%%%%%%%%%%%%%%%%%%%%%%%%%%%%%%%%%%%%%%%%%%%%%%%%%%%%%%%%%%%%%%%%%%%%%%

%%%%%%%%%%%%%%%%%%%%%%%%%%%%%%%%%%%%%%%%%%%%%%%%%%%%%%%%%%%%%%%%%%%%%%%%%%%%%%%%
% \section*{APPENDIX}

%\section*{ACKNOWLEDGMENT}

%%%%%%%%%%%%%%%%%%%%%%%%%%%%%%%%%%%%%%%%%%%%%%%%%%%%%%%%%%%%%%%%%%%%%%%%%%%%%%%%

\bibliographystyle{IEEEtran}
\bibliography{references}

\clearpage
\appendix
\normalsize

\subsection{Full List of Navigation Primitives}\label{sec:navigation-primitives}

% \oier{maybe add a table with all the primitives, or leave it for appendix}

Our full list of navigation primitives are listed in Table \ref{table:nav_primitives}.

\begin{table}[h]
  \centering
  \begin{tabularx}{0.45\textwidth}{l|X}
  \toprule

    primitives          & functions                       \\
\midrule
move\_to(pos)          & move to a position on the map.    \\
\rowcolor{light-gray}
move\_to\_left(object\_name)          & move to the left side of the nearest front object.    \\
move\_to\_right(object\_name)         & move to the right side of the nearest front object.    \\
\rowcolor{light-gray}
get\_pos(object\_name)                & get the map position of the nearest front object.    \\
get\_contour(object\_name)           & get the contour turning points of the nearest front object on the map.   \\
\rowcolor{light-gray}
with\_object\_on\_left(object\_name)  & turn until the nearest object is on the robot's left side.     \\
with\_object\_on\_right(object\_name)  & turn until the nearest object is on the robot's right side.  \\
\rowcolor{light-gray}
move\_in\_between(object\_a, object\_b) & move in between two objects. \\
turn(angle)           & turn right a certain angle. If the angle value is negative, turn left.  \\
\rowcolor{light-gray}
face(object\_name)           & turn until the nearest object is in front of the robot. \\
turn\_absolute(angle)           & turn to absolute angle. 0 is north, 90 is east, -90 is west, 180 is south. \\
\rowcolor{light-gray}
move\_north(object\_name)       & move to the north side of the nearest front object. \\
move\_south(object\_name)       & move to the south side of the nearest front object. \\
\rowcolor{light-gray}
move\_east(object\_name)       & move to the east side of the nearest front object. \\
move\_west(object\_name)       & move to the west side of the nearest front object. \\
\rowcolor{light-gray}
move\_to\_object(object\_name)       & move to the nearest object. \\
move\_forward(dist)       & move forward ``dist'' meters. \\

  \bottomrule
  \end{tabularx}
  \caption{List of the navigation primitives used.}
  \label{table:nav_primitives}
\end{table}

\subsection{Full Prompts}\label{sec:full-prompts}
Our full prompts used for getting the navigation results are listed below.
\lmp{
\prompt{
\# move a bit to the right of the fridge\\
robot.move\_to\_right(`refrigerator')\\
\# move in between the couch and bookshelf\\
robot.move\_in\_between(`couch', `bookshelf')\\
\# face the toilet\\
robot.face(`toilet')\\
\# move to the west of the chair\\
robot.move\_west(`chair')\\
\# turn right 20 degrees\\
robot.turn(20)\\
\# find any chairs in the environment\\
robot.move\_to\_object(`chair')\\
\# with the television on your left\\
robot.with\_object\_on\_left(`television')\\
\# move forward for 3 meters\\
robot.move\_forward(3)\\
\# move right 2 meters\\
robot.turn(90)\\
robot.move\_forward(2)\\
}
}

\lmp{
\prompt{
\# move back and forth to the chair and table 3 times\\
pos1 = robot.get\_pos(`chair')\\
pos2 = robot.get\_pos(`table')\\
for i in range(3):\\
\hspace*{4mm}robot.move\_to(pos1)\\
\hspace*{4mm}robot.move\_to(pos2)\\
\# move 3 meters south of the chair\\
robot.move\_south(`chair')\\
robot.face('chair')\\
robot.turn(180)\\
robot.move\_forward(3)\\
\# turn west\\
robot.turn\_absolute(-90)\\
\# turn east\\
robot.turn\_absolute(90)\\
\# turn south\\
robot.turn\_absolute(180)\\
\# turn north\\
robot.turn\_absolute(0)\\
\# turn east and then turn left 90 degrees \\
robot.turn\_absolute(90)\\
robot.turn(-90)\\
\# navigate to 3 meters right of the table\\
robot.move\_to\_right('table')\\
robot.face('table')\\
robot.turn(180)\\
robot.move\_forward(3)\\
}
}

\subsection{Prompt engineering.} For all methods in this work (including baselines), when using CLIP text encoding, instead of simply prompting the label of the object categories, we use the ensemble of prompt templates like ``A photo of {label}'', ``A picture of {label}'' mentioned in \cite{radford2021learning} to improve the retrieval performance. 

\subsection{Top-Down Map Semantic Segmentation}\label{sec:top_down_semantic}
For ablation purposes, we compute the semantic segmentation masks for the top-down maps in the Habitat simulator with the Matterport3D dataset. We use the collected RGB-D frames mentioned in Sec. \ref{sec:sim_experiments} to create the VLMaps and the CLIP on Wheels saliency maps. We evaluate all the semantic categories (the full list can be found in the link\footnote{\href{https://github.com/niessner/Matterport/blob/master/metadata/mpcat40.tsv}{https://github.com/niessner/Matterport/blob/master/metadata/mpcat40.tsv}}) supported in the Matterport3D dataset except ``void'', ``floor'', ``ceiling'', ``objects'', ``misc''. To get the ground truth semantic masks, we use the RGB-D frames and the ground truth image semantic masks to create a semantic top-down map. We back-project the depth pixels to the 3D space and project them to the top-down map. We assign the associated semantic values to the top-down map pixels. If multiple points are projected to the same location, we overwrite the old value if the new point's height is larger than the previous points. To compute semantic masks for VLMaps, we apply the open-vocabulary landmark indexing technique described in Sec. \ref{sec:sub_landmark_indexing} to the whole list of categories. To compute semantic masks for the CLIP on Wheels, we compute the saliency values and apply the same thresholding process as in \cite{gadre2022clip} to get a binary mask for each category. We evaluate the semantic segmentation metrics used in \cite{long2015fully}. The segmentation results is shown in Table \ref{table:seg_iou}.

We also show the IOU values of the top-10 frequent categories in Table \ref{table:seg_class}. The table shows that VLMaps performs better than CoW Map in most of the top-10 frequent categories. This is mainly because the GradCam used in CoW introduces a lot of noise in the saliency map, causing over-segmentation in the results. We also note that in the class ``seating'', VLMaps gets 0 IOU score. Since the LSeg model we used is pre-trained on segmentation datasets where some query classes might not be in the pre-defined training categories, LSeg's visual encoder will encode visually unseen objects (``seating'') to a similar seen object's embedding space (like ``chair'' or ``sofa'' here). As a result, visual-text misalignment could happen.

\begin{table}[h]
  \setlength\tabcolsep{5.2pt}
  \centering
  \begin{tabular}{lcccc}
  \toprule

    Metric              & CoW Map            & VLMaps (ours)                      \\
\midrule
pixel accuracy          & 66.1                  & \textbf{92.3}                                       \\
mean accuracy           & 9.6                  & \textbf{27.7}                                        \\
mIOU                    & 5.7                  & \textbf{19.0}                \\
frequency weighted mIOU & 42.9                  & \textbf{85.9}                \\
  \bottomrule
  \end{tabular}
  \caption{Top-Down Map Semantic Segmentation Results.}
  \label{table:seg_iou}
\end{table}

\begin{table}[h]
  \setlength\tabcolsep{5.2pt}
  \centering
  \begin{tabular}{lcccc}
  \toprule

Class    & Class Portion      & \multicolumn{2}{c}{IOU}       \\                
                                \cmidrule(lr){3-4}\
         &                    & CoW               & VLMaps (ours) \\
\midrule
wall     & 39.70              & 63.80             & \textbf{98.57}                      \\
chair    & 9.66               & 6.99              & \textbf{77.04}                      \\
table    & 8.14               & 1.20              & \textbf{13.82}                      \\
door     & 4.94               & 12.50             & \textbf{28.28}                      \\
seating  & 4.54               & \textbf{8.77}     & 0.00                                \\
stairs   & 3.75               & 12.35             & \textbf{22.02}                      \\
cabinet  & 3.38               & 1.13              & \textbf{1.87}                       \\
sofa     & 2.7                & 1.66              & \textbf{24.46}                      \\
bed      & 2.64               & 2.47              & \textbf{38.3}                       \\
shelving & 2.60               & 0.87              & \textbf{1.59}                       \\

  \bottomrule
  \end{tabular}
  \caption{Top-10 frequent per-class IOU}
  \label{table:seg_class}
\end{table}

\begin{figure*}[t]
    \centering
    \begin{subfigure}[t]{\textwidth}
      \centering
          \begin{subfigure}[t]{0.49\textwidth}
              \centering
              % include third image
              \includegraphics[height=7cm]{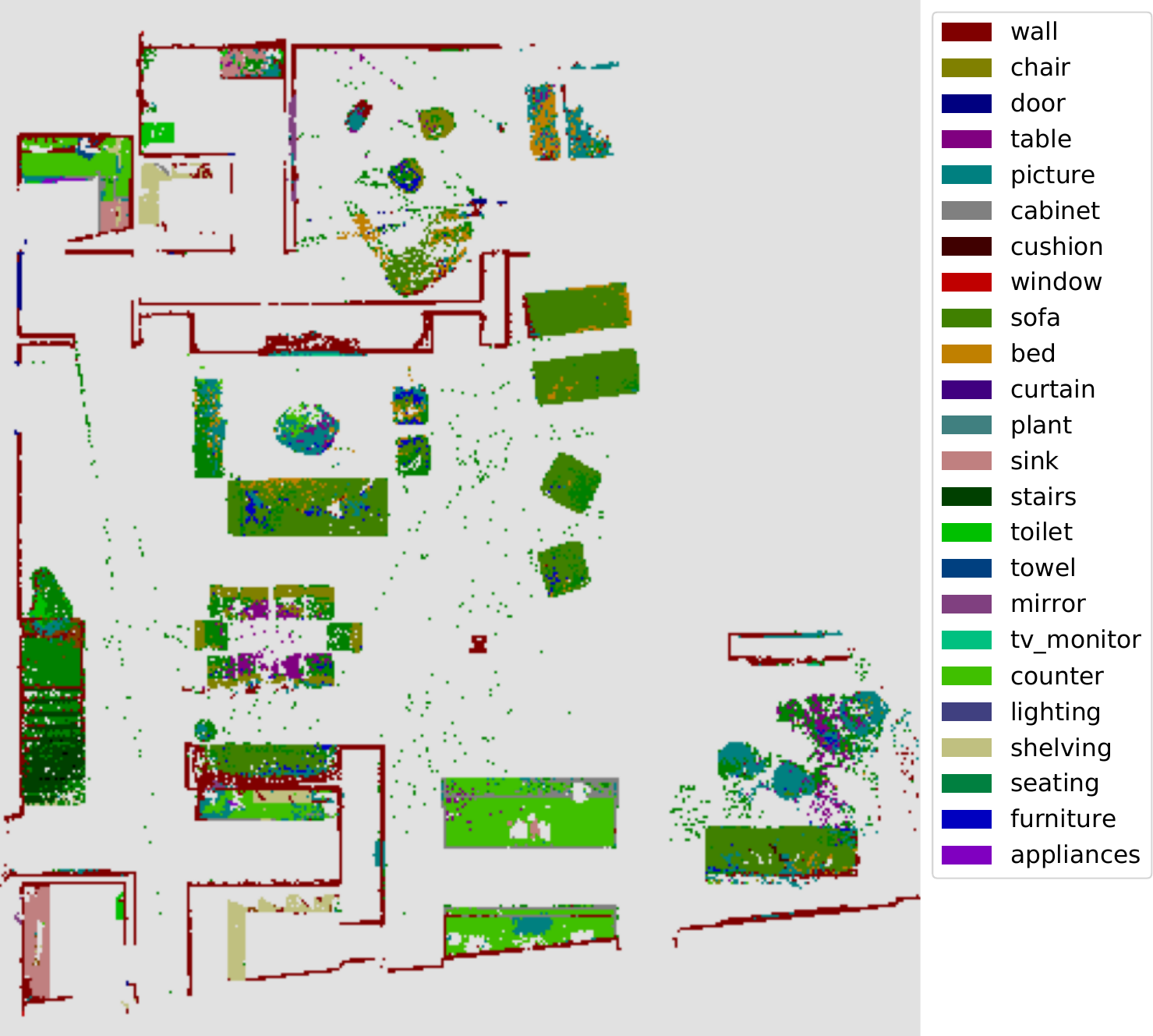}  
              \caption{VLMaps (sequence 1)}
              \label{fig:seg_vlmaps_1}
          \end{subfigure}
          \begin{subfigure}[t]{0.49\textwidth}
              \centering
              % include third image
              \includegraphics[height=7cm]{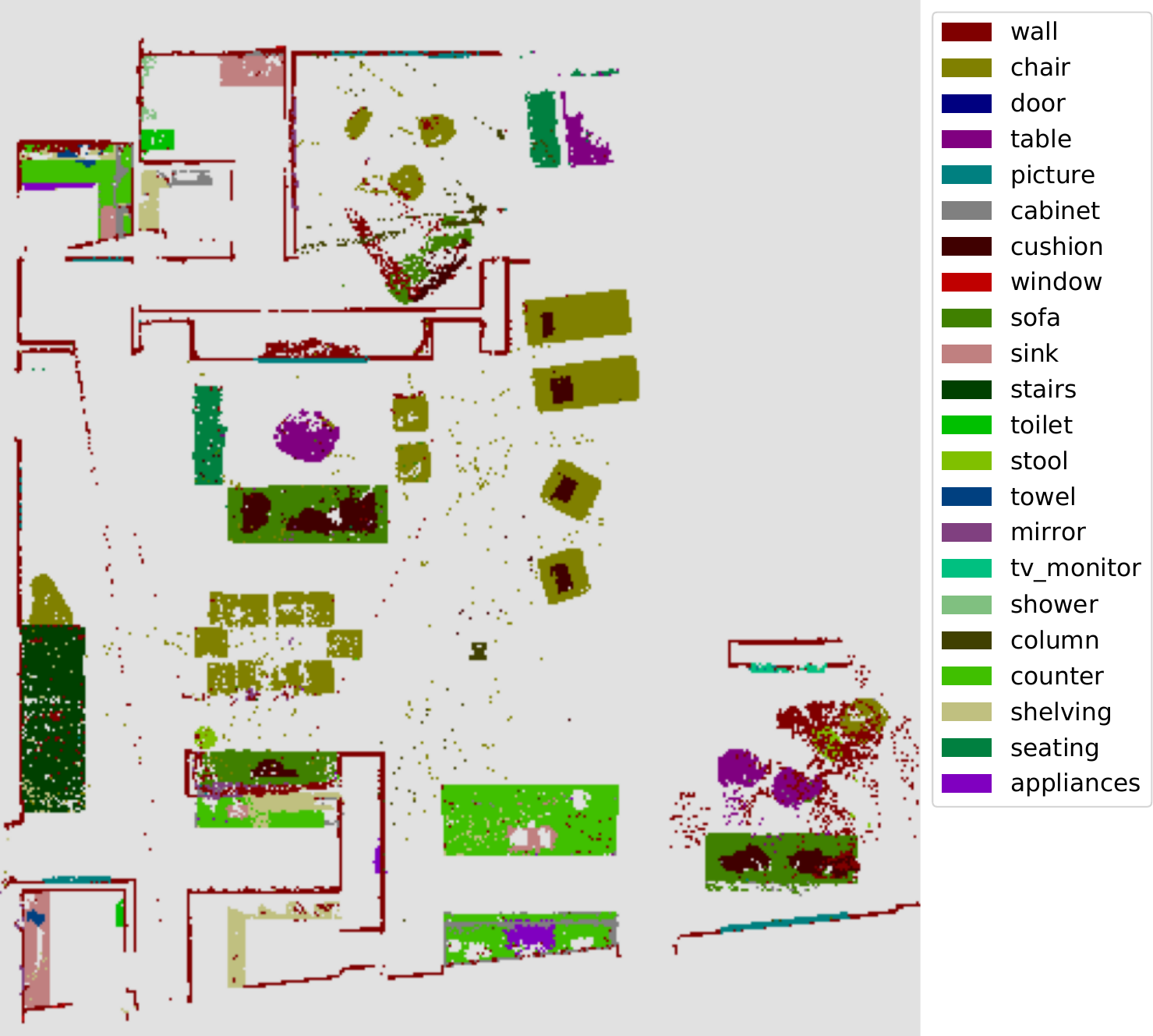}  
              \caption{GT (sequence 1)}
              \label{fig:seg_gt_1}
          \end{subfigure}

      \label{fig:seg_1}
    \end{subfigure}
    
    \begin{subfigure}[t]{\textwidth}
      \centering
          \begin{subfigure}[t]{0.49\textwidth}
              \centering
              % include third image
              \includegraphics[height=7cm]{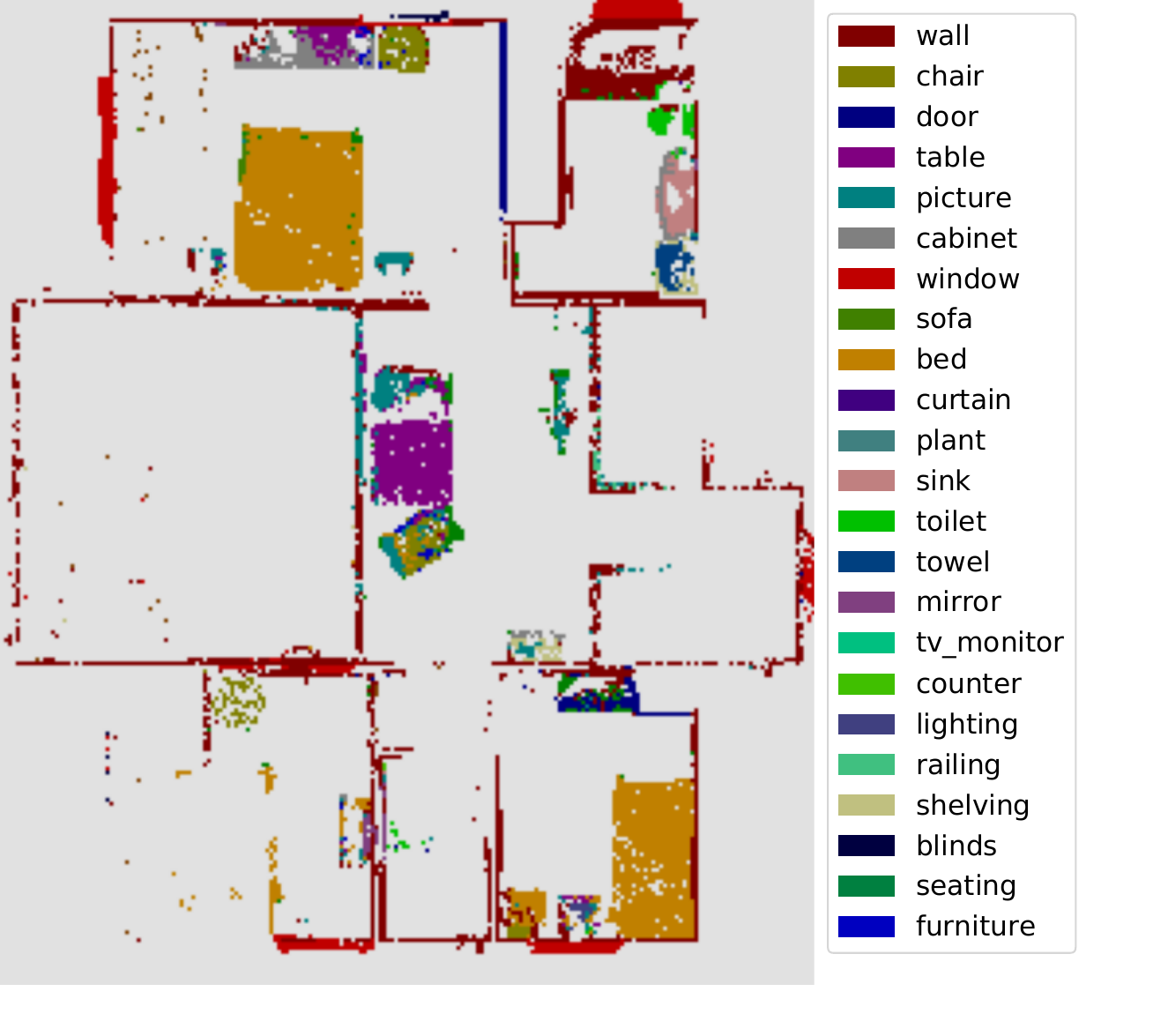}  
              \caption{VLMaps (sequence 2)}
              \label{fig:seg_vlmaps_2}
          \end{subfigure}
          \begin{subfigure}[t]{0.49\textwidth}
              \centering
              % include third image
              \includegraphics[height=7cm]{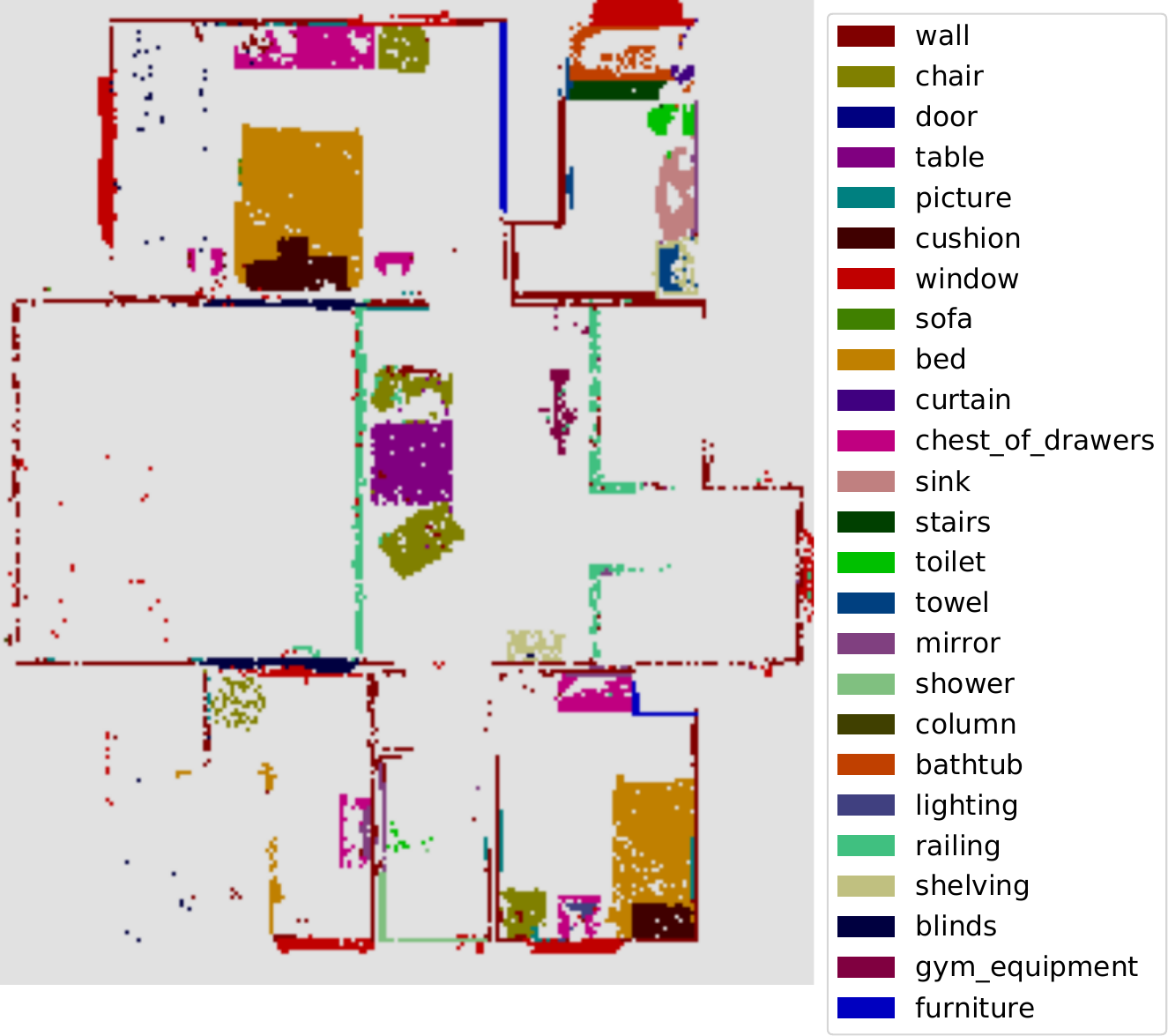}  
              \caption{GT (sequence 2)}
              \label{fig:seg_gt_2}
          \end{subfigure}
      \label{fig:seg_2}
    \end{subfigure}
    
    \begin{subfigure}[t]{\textwidth}
      \centering
          \begin{subfigure}[t]{0.49\textwidth}
              \centering
              % include third image
              \includegraphics[height=6cm]{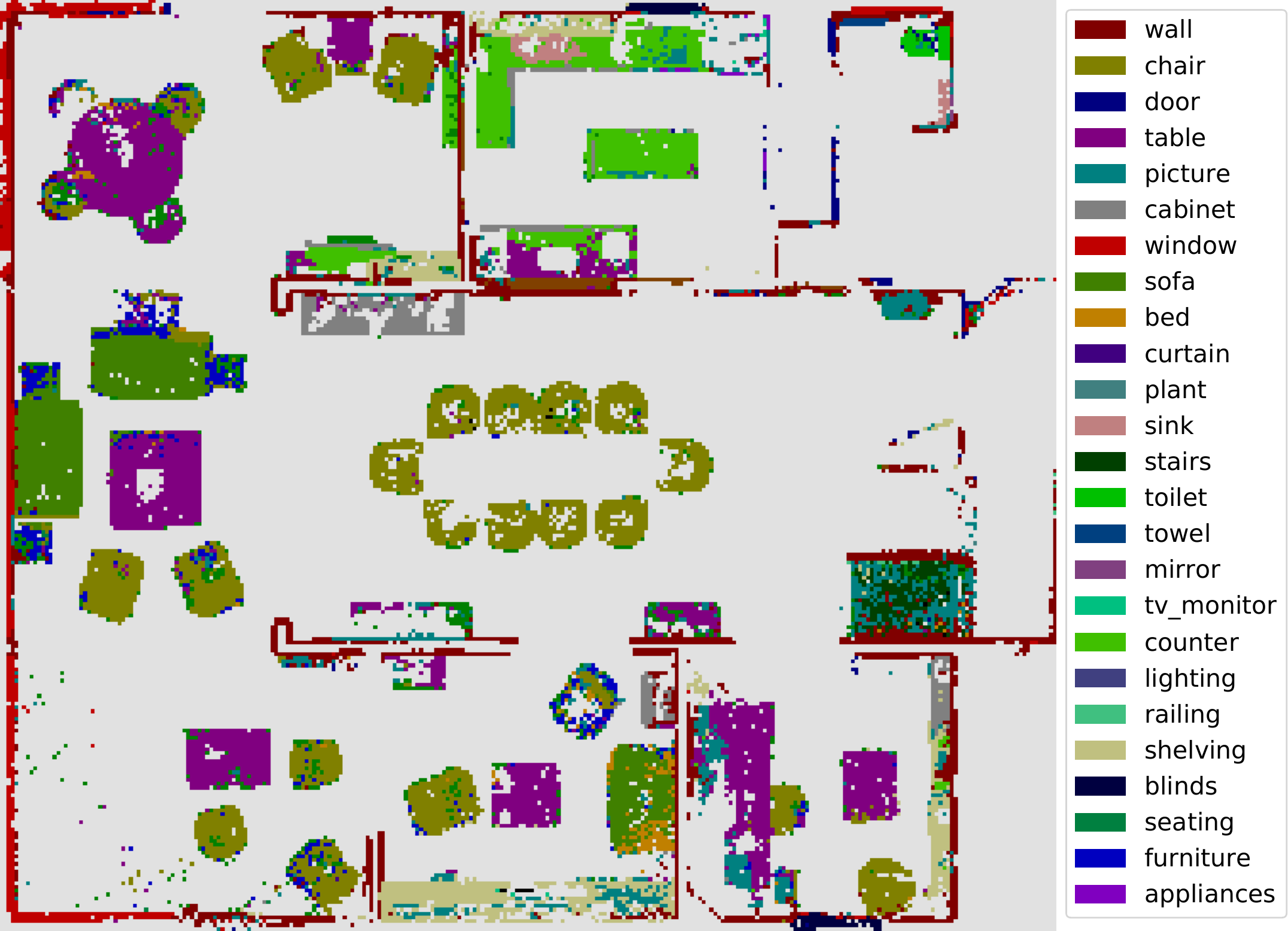}  
              \caption{VLMaps (sequence 3)}
              \label{fig:seg_vlmaps_3}
          \end{subfigure}
          \begin{subfigure}[t]{0.49\textwidth}
              \centering
              % include third image
              \includegraphics[height=6cm]{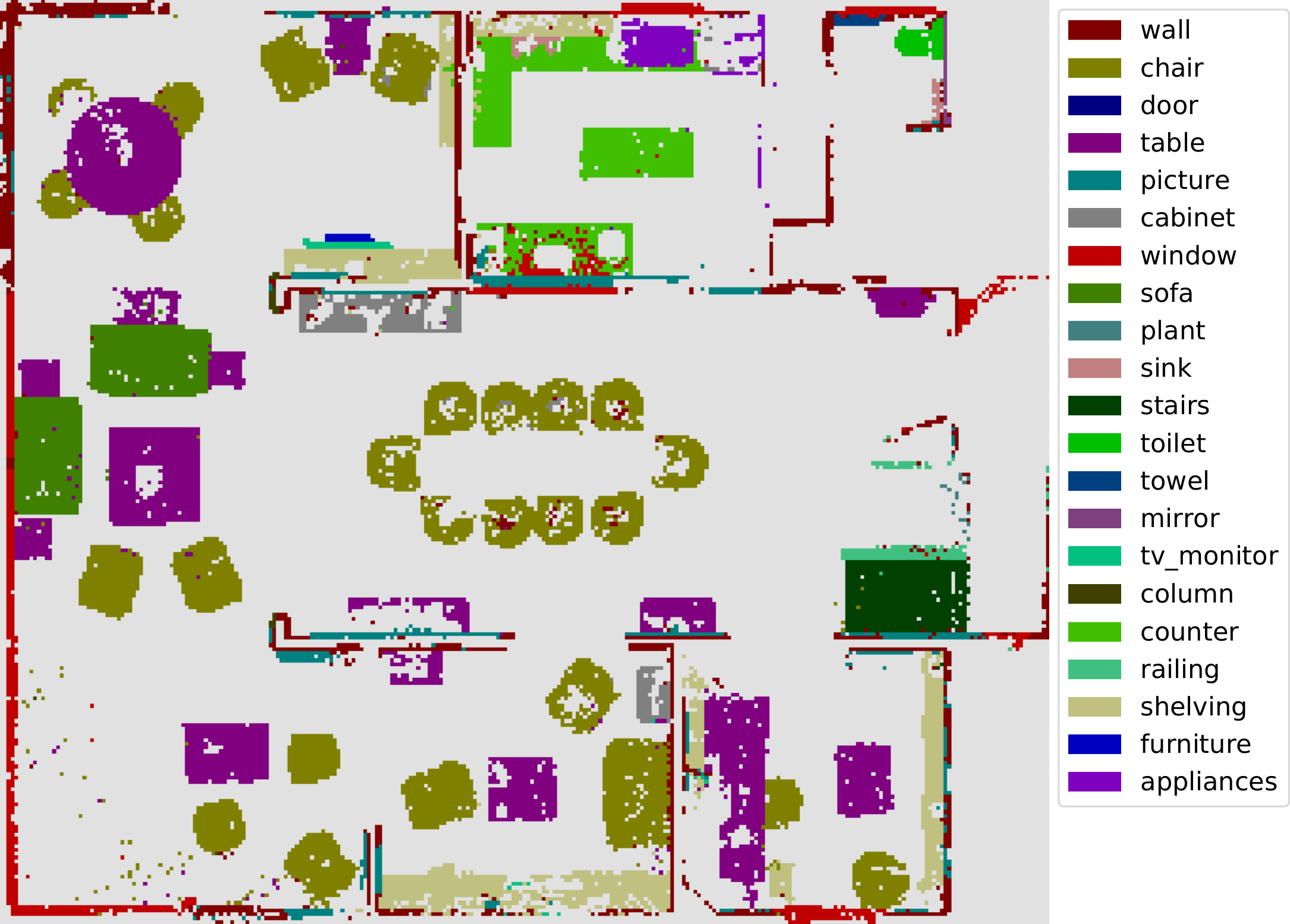}  
              \caption{GT (sequence 3)}
              \label{fig:seg_gt_3}
          \end{subfigure}
      \label{fig:seg_3}
    \end{subfigure}
	\caption{\small Qualitative semantic segmentation results}
	\label{fig:seg_qualitative}
	\vspace{-1em}
\end{figure*}

We visualize qualitative segmentation results in Figure \ref{fig:seg_qualitative}. We observe that for categories ``wall'', ``chair'', ``counter'', ``table'', and ``bed'', the segmentation results are mostly correct. Sometimes, when the ``sofa'' and ``chair'' are in similar material and shape (in Figure \ref{fig:seg_vlmaps_1} and \ref{fig:seg_gt_1}), VLMaps might fail to differentiate them, leading to wrong planning behaviors. We also observe from Figure \ref{fig:seg_vlmaps_2} and Figure \ref{fig:seg_vlmaps_3} that the segmentation of some objects are noisy. This could be caused by the features fusion strategy we adopt. For example, in the top left corner of Figure \ref{fig:seg_vlmaps_3}, there are some chairs and tables predictions with noise compared to the ground truth in Figure \ref{fig:seg_gt_3}. When we generate VLMaps for the scenes, we average the visual embeddings of points projecting to the same location on the top-down map. The averaging operation might introduce noise in the fused features, leading to noisy segmentation predictions (predicting ``sink'' on the table). In the future, more advanced fusion techniques can be explored to improve the segmentation results.

\subsection{Prompts for Obstacle Maps Generation}\label{sec:prompt_obs_map}
In Sec. \ref{sec:exp_multi_embodiment_navigation}, we generate open-vocabulary obstacle maps for a drone and a LoCoBot with the method introduced in Sec. \ref{sec:obstacle_maps}. For the LoCoBot (ground robot), we first define a potential obstacle list as [``chair'', ``wall'', ``wall above the door'', ``table'', ``window'', ``floor'', ``stairs'', ``other''] and perform open-vocabulary landmark indexing. Later, we only select the union of the masks for the objects ``wall'', ``chair'', ``table'', ``window'', ``stairs'', ``other'' as the obstacle map. For the drone (flying robot), we perform landmark indexing with the potential obstacle list: [``chair'', ``sofa'', ``wall'', ``table'', ``counter'', ``window'', ``floor'', ``stairs'', ``ceiling lights'', ``cabinet'', ``counter support'', ``other'']. Afterwards, we take union of the masks for [``wall'', ``window'', ``stairs'', ``ceiling lights'', ``cabinet'', ``other''] to generate the obstacle map.

\end{document}